\documentclass[a4paper]{article}

\usepackage{INTERSPEECH2020}
\usepackage{booktabs}
\usepackage{makecell}       % line breaks in tables
\usepackage{multicol}
\usepackage{multirow}
\usepackage{tikz}
\usetikzlibrary{calc}
\usetikzlibrary{positioning,arrows,automata}
\usetikzlibrary{shapes.geometric}
\usepackage{pgfplots}
\usepgfplotslibrary{groupplots}
\usepackage{subcaption}
\usepackage{amssymb}
\let\emptyset\varnothing
\usepackage{hyperref}

\definecolor{oCl}{RGB}{150,150,150}
\definecolor{Gray}{gray}{0.93}

\title{Low-Latency Sequence-to-Sequence Speech Recognition and Translation by Partial Hypothesis Selection}
\name{Danni Liu, Gerasimos Spanakis, Jan Niehues}
\address{Department of Data Science and Knowledge Engineering, Maastricht University, the Netherlands}
\email{\{danni.liu, jerry.spanakis, jan.niehues\}@maastrichtuniversity.nl}

\begin{document}

\maketitle
% https://www.overleaf.com/project/5e4fb0708d12ae00017d4441
\begin{abstract}
Encoder-decoder models provide a generic architecture for sequence-to-sequence tasks such as speech recognition and translation.
While offline systems are often evaluated on quality metrics like word error rates (WER) and BLEU scores, latency is also a crucial factor in many practical use-cases.
We propose three latency reduction techniques for chunk-based incremental inference and evaluate their accuracy-latency trade-off.
%Our best performing strategy utilizes the local stability of partial hypotheses as quality indicator.
On the 300-hour How2 dataset, we reduce latency by 83\% to 0.8 second by sacrificing 1\% WER (6\% rel.) compared to offline transcription. % this is baseline (bidirectional)
Although our experiments use the Transformer, the partial hypothesis selection strategies are applicable to other encoder-decoder models.
To reduce expensive re-computation as new chunks arrive, we propose to use a unidirectionally-attending encoder.
After an adaptation procedure to partial sequences, the unidirectional model performs on-par with the original model.
We further show that our approach is also applicable to speech translation.
On the How2 English-Portuguese speech translation dataset, we reduce latency to 0.7 second ($-$84\% rel.) while incurring a loss of 2.4 BLEU points (5\% rel.) compared to the offline system.
%The maximum number of pages is 5. The 5\textsuperscript{th} page may be used exclusively for references. The references should begin on an earlier page immediately after the Acknowledgements section, and continue onto the 5\textsuperscript{th} page. If no space is available on an earlier page, then the references may begin on the 5\textsuperscript{th} page.
%Index terms should be included as shown below.

\end{abstract}

\section{Introduction}
%\begin{itemize}
%\item However to fit tasks like live captioning: offline, latency, bad.
%\item How to handle? Structural modification (monotonic attention etc.) / Less intrusive ways.
%\item contributions.
%\end{itemize}
Encoder-decoder models with attention \cite{bahdanau2014neural, luong2015effective} opened the possibility of training many sequence tasks in an ``end-to-end'' fashion.
For automatic speech recognition (ASR), the roles of the acoustic and language models can be unified into one architecture \cite{chan2016listen, bahdanau2016end}.
Likewise, for speech translation, the traditionally cascaded modules for ASR and translation can be integrated \cite{liu2019end, sperber2019attention}.
In offline use-cases, promising results have been achieved, 
particularly with the recent adoption of self-attention mechanisms \cite{pham2019very, karita2019comparative, zeyer2019comparison, nguyen2019improving}.
However, latency still poses a challenge when applying encoder-decoder-attention models to online inference.
%The challenge is two-fold.
%From a computational point of view, as the decoder is autoregressive at test-time, constantly re-computing hidden states gets prohibitively expensive as more inputs stream in.
%From a modeling perspective, 
Under offline training, the decoder is conditioned on full encoded sequences via a soft attention distribution.
For incremental inference, when directly using a system trained on full-sequences, the train-test condition mismatch is likely to degrade performance.
Several approaches have been proposed to constrain the source-target attention distribution, notably monotonic attention \cite{raffel2017online, chiu2018monotonic, ma2020monotonic} and the Neural Transducer \cite{jaitly2016online, sainath2018improving}, where the training procedure is modified accordingly to account for possible alignments between input segments and output symbols.
% but there are tasks that are not monotonic, e.g. MT
% With the Transformer, multi-head attention make things more complicated (each head makes a different alignment)
%While these approaches require model-specific structural modifications, 

In this work, we take a different perspective: We start from a full-sequence model and study how to adjust its inference-time behavior for optimal latency-quality trade-off.
Moreover, we aim for a single model for both offline and online inference.
While the experiments use the Transformer \cite{vaswani2017attention}, our methods are agnostic to the underlying sequence representation mechanisms.
Based on the observation that small input chunks contain limited acoustic context, we eliminate unstable predictions by selectively outputting chunk-level hypotheses.
This partial hypothesis selection leads to an implicit look-ahead, as the model becomes more conservative with unstable predictions at chunk level.
To allow re-using decoder states as input streams in, we use an encoder with no dependency on future context.
To account for the different source-target attention distribution under incremental inference, we use a simple yet effective procedure to adapt to partial inputs.
%The experiments show 
The experiments show the effectiveness of our approach on ASR and speech translation.

%The experiments are based the Transformer model, as it has recently become competitive with recurrent neural networks (RNNs) on end-to-end ASR \cite{pham2019very, karita2019comparative, zeyer2019comparison, nguyen2019improving}. 
%However, our partial hypothesis selection strategies are generic to other encoder-decoder models.

\section{Related Work}
Recurrent neural network transducer \cite{graves2012sequence}, recurrent neural aligner \cite{Sak2017} and their self-attentional variants \cite{zhang2020transformer, dong2019self} are suitable to online end-to-end ASR as  they do not rely on global source-target attention.
% 1) RNN-T: combine CTC-based transcription network  + LM (prediction network), and has been successfully deployed in production \cite{he2019streaming}.
% 2) % without src-tgt attention, learns alignment
%These recently have been extended to Transformer-Transducer \cite{zhang2020transformer} and self-attention aligner \cite{dong2019self} respectively.  %A LATENCY-CONTROL END-TO-END MODEL FOR ASR USING SELF-ATTENTION NETWORK AND CHUNK-HOPPING (partially overlapping chunks)
For encoder-decoder-attention models, some form of input chunking is needed.
The alignment between chunks and output tokens is derived via external alignments \cite{novitasari2019sequence, inaguma2020minimum} or learned implicitly by connectionist temporal classification loss \cite{tian2020synchronous, miao2020transformer}.
To achieve adaptive chunk sizes, apart from monotonic chunkwise attention \cite{chiu2018monotonic}, triggered attention \cite{moritz2019triggered, moritz2020streaming} and adaptive computation time \cite{li2019end} are used.
In simultaneous text translation, the read-write decision of the decoder is modeled by fixed schedules \cite{cho2016can, ma2019simultaneous}, reinforcement learning agents \cite{gu2017learning, dalvi2018incremental}, or directly incorporated into the training objective \cite{arivazhagan2019monotonic}.
Re-translation is explored in conventional ASR systems \cite{nguyen2020low} and simultaneous (speech) translation \cite{niehues2018low, arivazhagan2020retranslation, arivazhagan2020retranslation2}.

\section{Incremental Decoding}
\iffalse
[shorten!!!]
The chunk-based decoding described above can be realized with Algorithm
1. In line 3, the number of input frames N and the fixed chunk length L together
determine the number of chunks in an utterance. In the loop between line 5 and
16, we operate on the input chunk by chunk. In practice, line 6 and 7 can be
further optimized. With the current algorithm, for each new chunk, the model
is re-decoding from the beginning of all inputs. The prefixes from earlier chunks
are then applied by forced decoding. In practice however, the re-decoding and
forced-decoding are not necessary, because some previously computed states
can be reused. For example, with the unidirectional encoder, we only need to
encode the current chunk, since the hidden representations from previous chunks
have no dependency on future context. In terms of the decoder, depending on
the specific behavior of the \textsc{prefix}($\cdot$) function, we can buffer decoder states differently to avoid re-computation.
\fi 
	\begin{figure}
	\scriptsize
	\tikzstyle{vector}=[rectangle,fill=none,draw=black,text=black,minimum width=0.5cm,minimum height=0.3cm,line width=0.5pt]
	\tikzstyle{io}=[circle,fill=white,draw=black,text=black,minimum width=0.5cm,minimum height=0.3cm]
	\tikzstyle{seq}=[rectangle,fill=none,draw=none,text=black,minimum width=1cm,minimum height=0.3cm]
	\tikzstyle{void}=[rectangle,fill=none,draw=none,text=black,minimum width=0.5cm,minimum height=0.3cm,inner sep=0,outer sep=0]
	\tikzstyle{att}=[->, draw=black]
	\tikzstyle{inp}=[->, draw=black]
	\tikzstyle{inpB}=[->,bend left=34, draw=black]
	\tikzstyle{recF}=[->, draw=black]
	\tikzstyle{recB}=[<-, draw=black]
	\definecolor{seqCol3}{rgb}{0.4,0.4,0.4}
	\definecolor{seqCol2}{rgb}{0.6,0.6,0.6}
	\definecolor{seqCol1}{rgb}{0.8,0.8,0.8}
	\centering
	\begin{tikzpicture}[>=stealth',shorten >=1pt,auto,node distance=0.7cm,
	triangle/.style = {regular polygon, regular polygon sides=3 }]
	
	\node[state][vector] at (-1.25,0) (e1) {Encoder};
	\node[state][vector] (e2) [below of=e1]   {Encoder};
	\node[state][vector] (e3) [below of=e2]   {Encoder};
	
	\node[state][vector] at (0.3,0) (d1) {Decoder};
	\node[state][vector] (d2) [below of=d1]   {Decoder};
	\node[state][vector] (d3) [below of=d2]   {Decoder};
	
	\node[state][seq,rounded corners=0.2cm, draw=black] at (2,0) (y1) {output$_1$};
	\node[state][seq,rounded corners=0.2cm, draw=black] (y2) [below of=y1]   {output$_2$};
	\node[state][seq,rounded corners=0.2cm, draw=black] (y3) [below of=y2]   {output$_3$};
	
	\node[state][void] at (1.15,-0.3) (p1) {\textsc{prefix}($\cdot$)};
	\node[state][void] (p2) [below of=p1]   {\textsc{prefix}($\cdot$)};
	\node[state][void] (p3) [below of=p2]   {\textsc{prefix}($\cdot$)};
	
	\node[state][seq, fill=seqCol1] at (-5,0) (seq1_1) {chunk$_1$};
	\node[state][seq, fill=none] at (-4.0,0) (seq2_1) {};
	\node[state][seq, fill=none] at (-3.0,0)  (seq3_1) {};
	
	\node[state][seq, fill=seqCol1]  (seq1_2) [below of=seq1_1] {chunk$_1$};
	\node[state][seq, fill=seqCol2,text=white]  (seq2_2) [below of=seq2_1] {chunk$_2$};
	\node[state][seq, fill=none]  (seq3_2) [below of=seq3_1] {};
	
	\node[state][seq, fill=seqCol1] (seq1_3) [below of=seq1_2]{chunk$_1$};
	\node[state][seq, fill=seqCol2,text=white] (seq2_3) [below of=seq2_2]{chunk$_2$};
	\node[state][seq, fill=seqCol3,text=white] (seq3_3) [below of=seq3_2]{chunk$_3$};

	\path  (seq3_1) edge[att] (e1) (e1) edge[att] (d1) (d1) edge[att] (y1)
	(seq3_2) edge[att] (e2) (e2) edge[att] (d2) (d2) edge[att] (y2)
	(seq3_3) edge[att] (e3) (e3) edge[att] (d3) (d3) edge[att] (y3);
	
	\draw[->,rounded corners] (y1) |- (p1);
	\draw[->,rounded corners] (y2) |- (p2);
	\draw[->,rounded corners] (y3) |- (p3);
	\draw[->,rounded corners] (p1) -| (d2);
	\draw[->,rounded corners] (p2) -| (d3);
	\draw[->,rounded corners] (p3) -| ++(-0.8,-0.2);
	\node[state][void] at (0.3,-1.95) (p4)   {$\dots$};
	
	\end{tikzpicture}
	\caption{Our incremental decoding framework, where an utterance is split into fixed-size chunks, and a subset of previous chunk-level outputs conditions the decoding of the next chunk.}
	\label{fig.chunkBasedSimulation}
\end{figure}
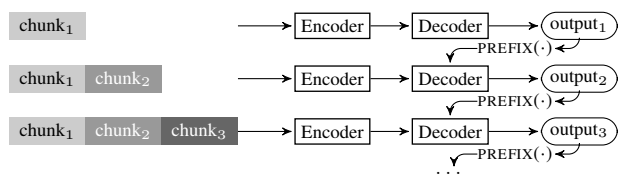
In offline systems, waiting till the end of input sequences is one of the largest factors contributing to latency.
For incremental inference, we divide the input utterance into fixed-size chunks and decode every time a new chunk arrives.
To avoid visual distractions from constantly modifying hypotheses, selected chunk-level predictions are \textit{committed to} and no longer modified.
Figure \ref{fig.chunkBasedSimulation} illustrates our framework. 
The decoding of the next chunk is conditioned by the predictions we have  \textit{committed to}.
In practice, the decoding of new chunks can either continue from a previously buffered decoder state, or start after forced decoding with the tokens we have committed to. 
In either approach, the source-target attention can span over all available chunks rather than only the current chunk.
% buffered decoder state.
% retransation.

%The chosen prefixes are committed to and no longer modified.
%When the next chunk arrives, its can only freely decode from the end of the previous hypotheses.
%By ``conditioning'', we refer to committing to a partial hypothesis and enforcing it as the basis for future outputs. 
%For example, in live captioning tasks, this means that the outputs displayed onscreen cannot be modified.

\subsection{Partial Hypothesis Selection} \label{subsec:partial.hyp}
%%% why take prefix.
Instead of outputting all chunk-level predictions, we are selective with the partial hypotheses for the following reasons.
First, the acoustic information towards chunk endpoints tends to be uncertain.
Moreover, early chunks often contain very limited context.
Therefore, some chunk-level predictions will be unstable, i.e. no longer part of the highest-likelihood hypothesis once new inputs appear.
As the decoder is autoregressive, ingesting these predictions can lead to further outputs of low quality.
In essence, by the partial hypothesis selection process, we intend to trade some latency for better output quality.
%To address this problem, we use a prefix function to select chunk-level outputs.

Formally, given chunk-level outputs $W^{(c)}$ from the $c$-th chunk, we only commit to $\textsc{prefix}(W^{(c)})$ and discard the rest.
This can be seen as a form of look-ahead, since the model becomes more conservative with outputting predictions.
The next chunk $c+1$ is conditioned by $\textsc{prefix}(W^{(c)})$ instead of the full $W^{(c)}$.
We describe several ways to realize the prefix function.
Depending on the behavior of the prefix function, we can buffer decoder states differently to avoid re-computation.

\textbf{Hold-$n$} One of the most straightforward ways to select partial hypotheses is to \textit{withhold} or \textit{delete} the last $n$ tokens in each chunk.
This gives the following prefix function:
\begin{equation} \label{eq.holdN}
\textsc{prefix}(W^{(c)}) = W_{0:\text{max}(0, |W ^ {(c)}| -n)}, \forall c \in \{1,\dots, C\},
\end{equation}
where $W^{(c)}$ is the output tokens from the $c$-th chunk, $|W ^ {(c)}|$ is the number of tokens in $W^{(c)}$, and $n$ is the number of tokens to be deleted.
In case $n$ is greater than the number of output tokens from the current chunk, the prefix will be empty, therefore displaying nothing for this chunk. 
On the other hand, hold-$0$ means fully outputting all chunk-level predictions.
%This is the reason for the tail index of $max(0, |W^{(c)}|-n)$ in Equation (\ref{eq.holdN}).

\textbf{Wait-$k$}
Inspired by techniques in simultaneous translation \cite{ma2019simultaneous}, we wait for the first $k$ chunks and subsequently outputs at a fixed rate $r$.
The prefix function $\textsc{prefix}(W^{(c)})$ is therefore: %distinguishes between two cases, with the first one corresponding to waiting, and the second to outputting at rate $r$:
\begin{equation} \label{eq.waitK}
\begin{cases}
\emptyset,& \text{if } c \le k \\
W^{(c)}_{0 : \text{min}(|W ^ {(c)}|, r)}, & \text{otherwise,} 
\end{cases}
\end{equation}
where $\text{min}(|W^{(c)}|, r)$ ensures that when the output rate $r$ exceeds the number of predicted tokens, all chunk-level outputs are shown. In this case, the effect is identical to hold-$0$.

\textbf{Local Agreement}
%Our goal is to output promising partial hypotheses promptly.
If a partial hypothesis is unlikely to change when new inputs arrive, it can be considered promising.
This idea is central to the partial track-back procedure in conventional ASR systems \cite{partial1982brown}.
Here, we realize this idea by considering local agreement, and taking the agreeing prefixes of two consecutive chunks as stable hypotheses.
For the first chunk, we do not display any output since there is not yet a previous chunk to compare with.
From the second chunk onwards, we seek agreement with the predecessor's outputs.
We take the longest common prefix of the current chunk's outputs $W^{(c)}$ and the not-yet-displayed outputs of the previous chunk. 
Let $\textsc{discard}(\cdot)$ indicate the tokens that are not displayed, i.e. those in $W^{(c)}$ but not in $\textsc{prefix}(W^{(c)})$.
The behavior of the local agreement strategy is defined as $\textsc{prefix}(W^{(c)})$=
\begin{equation}
\begin{cases}
\emptyset,& \text{if } c = 1\\
\textsc{LCP}(\textsc{discard}(W^{(c-1)}), W^{(c)}), & \text{otherwise,}
\end{cases}
\end{equation}
where \textsc{LCP($\cdot$)} indicates longest common prefix of two lists.

\subsection{Unidirectional Encoder}
To facilitate incremental inference, we train a Transformer with unidirectional encoder.
By masking future context, we let the encoder only attend to previous and current positions. 
This is analogous to using unidirectional RNN encoders in recurrent models.
Since each position has no future dependency, the existing decoder hidden states also do not need to be re-computed as new input chunks arrive.
Note that the unidirectional model is still trained with full utterances.
A later adaptation step (Sec. 4) addresses partial inputs under incremental decoding. 

\subsection{Latency Measurement}
In the strictest sense, the latency of transcribing a word is the elapsed time between when it is said and when the corresponding transcription is generated.
Therefore, for an output sequence $w_{1, \dots, T}$, the average latency is
\begin{equation} \label{eq.latency1}
\frac{1}{T} {{\sum}}_{t=1}^{T}(\textrm{outputTime}(w_t)- \textrm{endTime}(w_t)),
\end{equation}
which can be rewritten to
\begin{equation} \label{eq.latency2}
\frac{1}{T} {{\sum}}_{t=1}^{T}\textrm{outputTime}(w_t) - \frac{1}{T} {{\sum}}_{t=1}^{T} \textrm{endTime}(w_t).
\end{equation}
However, the second term in Equation (\ref{eq.latency2}) is usually unknown since the transcriptions are not time-aligned.
Alternatively, assuming all output words correspond to the ground truth, the second term in Equation (\ref{eq.latency2}) remains constant and can be dropped when comparing the latency of two systems, therefore resulting in a measurement only based on the first term.
%\begin{equation} \label{eq.latency3}
%\frac{1}{T} {{\sum}}_{t=1}^{T}\textrm{outputTime}(w_t).
%\end{equation}
Note that this metric is only valid for latency differences.
When considered in isolation, it does not correspond the actual latency experienced by users. %since it assumes that all words are said at the beginning of an utterances.
In the framework of chunk-based decoding, the output timestamps is easily derived.
Since the words are displayed at the boundaries of fixed-size chunks, the output timestamp is the product of the relevant chunk index and the chunk length.

\section{Adaptation to Partial Inputs}
\subsection{Train-Test Condition Mismatch}
\begin{figure}[h]
	\normalsize
	\pgfplotsset{heat left/.style={
			axis line style={white},
			point meta max=0.5,
			point meta min=0,
			x grid style={white},
			xmin=0, xmax=196,
			xtick style={color=white!15!black},
			xtick={0.5,32.5,64.5,96.5,128.5,160.5,192.5},
			xticklabels={0,32,64,96,128,160,192},
			y dir=reverse,
			y grid style={white},
			ymin=0.5, ymax=23.5,
			ytick style={color=white!15!black},
			ytick={0.5,1.5,2.5,3.5,4.5,5.5,6.5,7.5,8.5,9.5,10.5,11.5,12.5,13.5,14.5,15.5,16.5,17.5,18.5,19.5,20.5,21.5,22.5,23.5},
			yticklabels={we,'re,going,to,work,on,a,arm,drill,that,will,help,you,have,gr\_,ace\_,ful,hand,movements,in,front,of,you,.}
		}
	}
	\pgfplotsset{heat right/.style={
			axis line style={white},
			colorbar,
			colorbar style={ytick={0,0.1,0.2,0.3,0.4,0.5},yticklabels={0.0,0.1,0.2,0.3,0.4,0.5},ylabel={}},
			colormap={mymap}{[1pt]
				rgb(0pt)=(0.9922,0.9843,0.9843);
				rgb(3pt)=(0.987837818696884,0.9733,0.972544759206799);
				rgb(35pt)=(0.952716666666667,0.8816,0.8779);
				rgb(43pt)=(0.943992304060434,0.8635,0.854389518413598);
				rgb(58pt)=(0.926431728045326,0.8275,0.807067138810198);
				rgb(101pt)=(0.878112181303116,0.743,0.676855240793201);
				rgb(105pt)=(0.87375,0.73765,0.6651);
				rgb(109pt)=(0.869387818696884,0.7323,0.655545355191257);
				rgb(117pt)=(0.860663456090651,0.7216,0.63643606557377);
				rgb(125pt)=(0.851827242681775,0.7103,0.617081785063752);
				rgb(133pt)=(0.843102880075543,0.6991,0.597972495446266);
				rgb(184pt)=(0.786058970727101,0.6549,0.473027140255009);
				rgb(192pt)=(0.777222757318225,0.6481,0.453672859744991);
				rgb(199pt)=(0.768498394711993,0.6447,0.434563570127505);
				rgb(207pt)=(0.75977403210576,0.6413,0.415454280510018);
				rgb(211pt)=(0.7553,0.6392,0.405654644808743);
				rgb(215pt)=(0.753669955654102,0.642568181818182,0.3961);
				rgb(302pt)=(0.7176,0.7171,0.368824581939799);
				rgb(450pt)=(0.4719,0.651722448979592,0.321732608695652);
				rgb(454pt)=(0.463230931263858,0.650011224489796,0.3205);
				rgb(478pt)=(0.410994235033259,0.6397,0.306801658163265);
				rgb(533pt)=(0.288960421286031,0.608095777591973,0.2748);
				rgb(541pt)=(0.2714,0.603547993311037,0.284470320130222);
				rgb(717pt)=(0.1882,0.502,0.5004);
				rgb(792pt)=(0.138484277620397,0.3841,0.486876892430279);
				rgb(835pt)=(0.10976111898017,0.3035,0.479063944223108);
				rgb(890pt)=(0.0731740793201133,0.1893,0.469111952191235);
				rgb(968pt)=(0.0209259206798867,0.00897047387606321,0.4549);
				rgb(972pt)=(0.0183268413597734,0,0.4329);
				rgb(992pt)=(0.00519815864022665,0,0.2212);
				rgb(996pt)=(0.00259907932011332,0,0.1684);
				rgb(1000pt)=(0,0,0)
			},
			point meta max=0.5,
			point meta min=0,
			x grid style={white},
			xmin=0, xmax=196,
			xtick style={color=white!15!black},
			xtick={0.5,32.5,64.5,96.5,128.5,160.5,192.5},
			xticklabels={0,32,64,96,128,160,192},
			y dir=reverse,
			y grid style={white},
			ymajorticks=false,
			ymin=0.5, ymax=23.5,
			ytick style={color=white!15!black},
			ytick={0.5,1.5,2.5,3.5,4.5,5.5,6.5,7.5,8.5,9.5,10.5,11.5,12.5,13.5,14.5,15.5,16.5,17.5,18.5,19.5,20.5,21.5,22.5,23.5},
			yticklabels={we,'re,going,to,work,on,a,arm,drill,that,will,help,you,have,gr\_,ace\_,ful,hand,movements,in,front,of,you,.}
		}
	}
	\centering
	\resizebox{0.45\textwidth}{!}{
		\begin{tikzpicture}
			\begin{axis}[heat left]
				\addplot graphics [includegraphics cmd=\pgfimage,xmin=0, xmax=196, ymin=23.5, ymax=0.5] {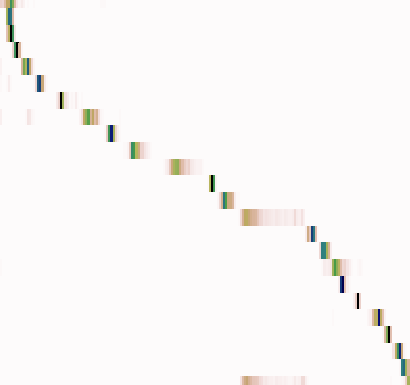};
			\end{axis}
		\end{tikzpicture}
		\begin{tikzpicture}
			\begin{axis}[heat right]
				\addplot graphics [includegraphics cmd=\pgfimage,xmin=0, xmax=196, ymin=23.5, ymax=0.5] {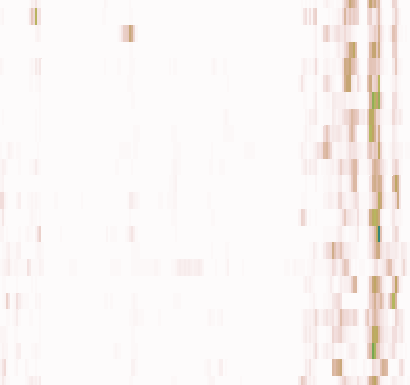};
			\end{axis}
		\end{tikzpicture}
	}
\caption{Example source-target attention weights at first decoder layer of our unidirectional-encoder model. The vertical axis goes top-down and corresponds to decoding the tokens sequentially. The horizontal axis indicates the input frames.}
\label{fig.attWeightsUni}
\end{figure}

When training an offline system, the model is rarely incentivized to generate incomplete sentences, since most reference transcriptions are full sentences.
Indeed, offline systems trained on full sequences were found to fantasize full outputs even when given partial inputs \cite{niehues2018low}.
More importantly, under chunk-based incremental decoding, the decoder has to operate on partial encoder representations.
During training, although the encoder has no future context at each time step, the decoder can still access the full encoded sequences.
Therefore, the source-target attention can compensate for the loss of future context by placing heavier weights towards the end of input sequences.
%The division of labor among attention heads has \cite{tsunoo2019online}.
In Figure \ref{fig.attWeightsUni}, we show an example from our experiments with unidirectional encoder, where one attention head aligns input frames with output tokens, while the other focuses on the tail area of the inputs, thereby compensating for the lack of future context in the encoder representation.
This showcases the train-test condition mismatch that needs to be resolved.
%This way, despite the encoded hidden representation being unidirectional and lacking future context, the source target attention is still able to compensate for this 
%%%%%%%%%%%%%%%%%%%%%%%%%%%%

\subsection{Fine-Tuning with Partial Inputs} \label{subsec:finetune}
To bridge the mismatch between train and test conditions, we adapt the trained models using partial sequences.
%Adapting a trained neural network to a new task is prone to catastrophic forgetting, a phenomenon where the network forgets the learned knowledge and loses performance on the previous task. 
To avoid losing performance on offline inference after adaptation, we take inspiration from the approach used in \cite{niehues2018low} and continue training on a 1:1 mix of both full and partial sequences. 
In absence of time-aligned transcriptions, we make an approximation by taking proportionally many input frames and output tokens.
Specifically, given audio frames $U_{1, \dots, I}$, transcription $T_{1, \dots, J}$, and ratio $p$, we take the first $\lceil I \cdot p \rceil$ frames and $\lceil J \cdot p \rceil$ tokens.
We expect the fine-tuning step to primarily adjust source-target attention distribution to work with both full and partial inputs.

\section{Experimental Setup}
\subsection{Data}
We use the 300-hour How2 dataset \cite{sanabria18how2}, which contains instruction videos from YouTube.
For transcription and translation, we use the English audio with English and Portuguese subtitles respectively.
We use the pre-complied 40-dimensional filter bank, and exclude three dimensions of pitch information, the calculation of which involves future frames.
To account for online inference, we do not use utterance or speaker-level normalization.
An overview of the ASR dataset is show in Table \ref{tab.corpusStats}.

%, TED-LIUM v3 \cite{hernandez2018ted}, preprocessing using Kaldi (no normalization) \cite{povey2011kaldi}

\begin{table}[h]
	\footnotesize
	\centering
	\caption{Overall statistics of the How2 ASR dataset.}
	\begin{tabular}{rrrrrr}
		\toprule
		\textbf{Dataset} & 
		\textbf{Len (h:m)} & 
		\textbf{Total utt.} & 
		\textbf{Total words} & 
		\textbf{Avg utt. len}\\
%		\textbf{\makecell[tl]{Language}} & 
%		\textbf{\makecell[tl]{Total \\ length \\ (h:m)}} &
%		\textbf{\makecell[tl]{Total \\ utterances}} & 
%		\textbf{\makecell[tl]{Total \\ words}} & 
%		\textbf{\makecell[tl]{Average \\ utterance \\ length (s)}} \\
		\midrule
%		\multicolumn{1}{l}{How2}  \\
		train & 298:12 & 184,949 & 3,304,534 &  5.80 \\
		dev   &   3:15 &   2,022 &    36,013  &  5.78 \\
		test  &   3:43 &   2,305 &    40,890  &  5.80 \\
		%\midrule
%		\multicolumn{1}{l}{TED} \\
%		train & 452:04 & 268,262 & 4,554,246 &  6.07 \\
%		dev   &   1:36 &     507 &    17,783 & 11.36 \\
%		test  &   2:37 &   1,155 &    27,500 &  8.16 \\
		\bottomrule
	\end{tabular}
	\label{tab.corpusStats}
\end{table}

\subsection{Hyperparameters}
When training the Transformer model\footnote{Implementation and experiment recipes available at: \newline https://github.com/dannigt/NMTGMinor.lowLatency}, we follow the reported hyperparameters from \cite{pham2019very}, including the optimizer choice, learning rate, warmup
steps, dropout rate, label smoothing rate, and embedding dimension.
Several hyperparameters differ. 
The size of the inner feed forward
layer is 2048. 
We use 32 encoder and 12 decoder layers, and byte-pair-encoding \cite{sennrich2016neural} of size 10,000.
%The number of sequences per batch depends on GPU memory, but we limit each batch to have a maximum of 2048 words or 128 sentences.
The final model is an ensemble of the last 5 best checkpoints.
We use a beam width of 8 when decoding.

\subsection{Adaptation Procedure}
For each training utterance, we choose a partial transcription from 10\% to 40\% of the original number of tokens. 
The ratio is intentionally kept low to create partial inputs lacking future information.
Based on the chosen ratio, we proportionately take a subset of the input audio frame as specified in Section \ref{subsec:finetune}.
Then we mix the partial sequences and their transcriptions with the original training instances. 
The learning rate is reduced to a quarter of before. 
Moreover, we use the original full-sequence dev set to avoid losing performance on offline inference.

\section{Experiments and Results}
In the following experiments, we use a fixed chunk size of 0.5 second. Whiles we experimented with other chunk size (1 and 2 seconds), the key observations do not differ from those presented here.
When decoding for new chunks, the partial hypotheses we have committed to are fed in via forced decoding.
%\input{tabs/offline}
%Table \ref{tab.offlineWER}
\subsection{Bounds for Accuracy and Latency}
We first consider the upper and lower bounds for accuracy and latency.
Offline transcription is expected to yield the lowest error rate but also the highest latency.
On the contrary, when all chunk-level hypotheses are immediately shown (the hold-$0$ strategy), we expect the lowest latency at the cost of compromising accuracy.
Therefore, for incremental decoding, the word error rate is bounded between that of offline and hold-$0$, whereas latency is bounded between that of hold-$0$ and offline.
%WER between hold-0 and offline, latency between offline and hold-0.
%since we do not consider correcting past outputs while online decoding, 
%These two will be bounds for the coming experiments. \textbf{we report ... relative to ...}
%Nevertheless, since no short utterance has been seen in training, it is possible that, especially in earlier short chunks, the model generates fictitious outputs that do not exist in audio inputs. 
%This would cause longer output sequences on average. 
%As our latency metric is partially based on the number of output tokens, these excessive tokens in earlier chunks could skew the calculated latency to be lower than actual. Therefore, it must be noted that the hold-0 latency should not be treated as a strict lower bound.

%%%%%%%%%%%%%%%%%%%%%%%%%%%%%%%%%%%%%%%%%%%%%%%%%%%%%%%%%%%%%%%%%%%%%%%%%%%%%%%
\subsection{Accuracy-Latency Trade-Off of Different Strategies}
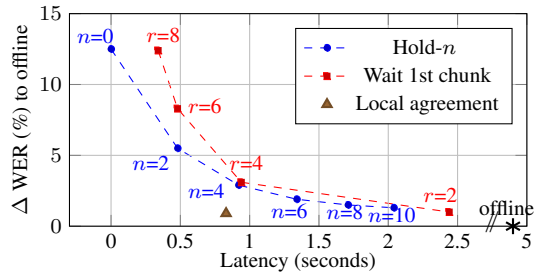
\begin{figure}[h]
	\centering
	\footnotesize
	\begin{tikzpicture}
	\pgfplotsset{
		scaled y ticks = false,
		width=0.95\columnwidth,
		height=0.55\columnwidth,
		axis on top,
		grid=major,
		xlabel style={at={(0.5,0.1)}},
		extra x tick style={grid=none, draw=none, tick label style={xshift=0cm,yshift=.30cm},tick style={draw=none}},
		extra x tick label={$\mathbin{/\mkern-6mu/}$},
		ylabel shift={-1.5em},
		ylabel style={align=center, at={(0.1,0.5)}}
%		try min ticks=10
	}
	\begin{groupplot}[ 
	group style={
		group size=1 by 1,
		vertical sep=0pt,
		horizontal sep=0pt
	},
	]
	% ----------
	% Plot [0, 0]
	%-----------
	\nextgroupplot[
	xlabel={Latency (seconds)},
	ylabel={$\Delta$ WER (\%) to offline},
	xticklabels={0,0.5,1,1.5,2,2.5,5}, % Fake here the x-value for the offline thing
	xtick={0,0.5,1,1.5,2,2.5,3},
	extra x ticks={2.75},
	xmin=-0.3,
	xmax=3,
	ymin=0, %13.9,
	ymax=15,
	legend style={legend columns=1,fill=white,draw=black,anchor=west,align=left, font=\footnotesize}, %at={($(0,0)+(1cm,1cm)$)},
	legend to name=legend1
	]
	\addplot+[sharp plot, dashed, mark size=1.3pt] table {
		x       y
		0       12.5
		0.48    5.5
		0.92    2.9
		1.34    1.9
		1.71	1.5
		2.04	1.3
	};
	\addplot+[sharp plot, dashed, mark size=1.3pt] table {
		x       y
		0.34    12.4
		0.48    8.3
		0.94    3.1
		2.44    1.0
	};
	% agree
	\addplot+[sharp plot, only marks,mark=triangle*,thick,mark size=2pt] table {
		x       y
		0.83  0.9
	};
	\addplot+[sharp plot, only marks,mark=asterisk,thick,mark size=3pt] table {
		x       y
		2.9    0 
	}; % was 14.9
	\iffalse
	\addplot[
	black,
	only marks,
	mark=none,
	visualization depends on=\thisrow{alignment} \as \alignment,
	nodes near coords,
	point meta=explicit symbolic,
	every node near coord/.style={anchor=\alignment} 
	] table [
	meta index=2 
	] {
		x       y         label       alignment
		-0.1  33.4    $n$=0       180
		1       21.5    $n$=2       180
		1.8    18.9    $n$=4       180
		2.6    18.0    $n$=6       180
	};
	\fi 
	
	\addlegendentry{Hold-$n$    };    
	\addlegendentry{Wait 1st chunk};    
	\addlegendentry{Local agreement}; 
%	\addlegendentry{Offline};    

	% (Relative) Coordinate at top of the first plot
	\coordinate (c1) at (rel axis cs:0,1);
	%-----------
	% Plot [0, 1]
	%-----------
	\iffalse
	\nextgroupplot[
	title={Unidirectional},
	xlabel={Latency (seconds)},
	ylabel={},
	xticklabels={0,0.5,1,1.5,2,2.5,5}, % Fake here the x-value for the offline thing
	xtick={0,0.5,1,1.5,2,2.5,3},
	extra x ticks={2.75},  
	xmin=-0.3,
	xmax=3,
	ymin=0,%13.9,
	ymax=28,
	]
	\addplot+[sharp plot, dashed, mark size=1.2pt] table {
		x       y
		%	    	0      40.7
		0.42   22.0
		0.83   18.8
		1.22   17.6
		1.58   17.2
		1.92   16.8
	};
	
	\addplot+[sharp plot, dashed, mark size=1.2pt] table {
		x       y
		0.68	25.7
		0.71	22.9
		1.04	18.5
		2.37	15.4
	};
	% agree
	\addplot+[sharp plot, only marks,mark=triangle*,thick,mark size=1pt] table {
		x       y
		0.54  16.8
	};
	% optimal
	\addplot+[sharp plot, only marks,mark=asterisk,thick,mark size=2pt] table {
		x       y
		2.9      14.5
	};
	\fi

	\addplot[
	blue,
	only marks,
	mark=none,
	mark size=0.5pt,
	visualization depends on=\thisrow{alignment} \as \alignment,
	nodes near coords,
	point meta=explicit symbolic,
	every node near coord/.style={anchor=\alignment} 
	] table [
	meta index=2 
	] {
		x       y         label       alignment
		-0.3   13.4    $n$=0       180
		0.1    4.5    $n$=2       180
		0.5    2.5    $n$=4       180
		1.1    1.1    $n$=6       180
		1.5    1.0    $n$=8       180
		1.8    0.8    $n$=10       180
	};
	\addplot[
	red,
	only marks,
	mark=none,
	mark size=0.5pt,
	visualization depends on=\thisrow{alignment} \as \alignment,
	nodes near coords,
	point meta=explicit symbolic,
	every node near coord/.style={anchor=\alignment} 
	] table [
	meta index=2 
	] {
		x       y         label       alignment
		0.2    13.5    $r$=8       180
		0.5    8.5    $r$=6       180
		0.8    4.5    $r$=4       180
		2.2    2.2    $r$=2       180
	};
	\addplot[
	black,
	only marks,
	mark=none,
	visualization depends on=\thisrow{alignment} \as \alignment,
	nodes near coords,
	point meta=explicit symbolic,
	every node near coord/.style={anchor=\alignment} 
	] table [
	meta index=2 
	] {
		x       y         label       alignment
		2.6    1.3    offline       180
	};
	% (Relative) Coordinate at top of the second plot
%	\coordinate (c2) at (rel axis cs:1,1);
	\end{groupplot}
%	\coordinate (c3) at ($(c1)!.5!(c2)$);
	\node[above] at (4.4,1.25)
	{\pgfplotslegendfromname{legend1}};
	\end{tikzpicture}%
	\caption{Accuracy-latency trade-off of different partial hypothesis selection strategies. $n$ indicates the number of removed last tokens. $r$ indicates the fixed output rate (tokens/second).}
	\label{fig.tradeoffHow2}
\end{figure}

%\begin{itemize}
%\item really don't have to wait until the end!
%\item all the strategies work 
%\item wait-k not good! kick out.
%\end{itemize}
%\input{tabs/beforeAfterAdapt}
%Table \ref{tab.full}: hold-0 and offline
First, we evaluate the hypothesis selection strategies proposed in Section \ref{subsec:partial.hyp} in terms of accuracy-latency trade-off. 
Here we use the full-sequence model with the original bidirectional encoder.
The results are summarized in Figure \ref{fig.tradeoffHow2} for visual clarity.
The horizontal axis indicates the latency difference to the hold-$0$ strategy, the lower bound for latency.
The vertical axis indicates the WER difference to the offline system, the lower bound for accuracy.
The data points on the interpolated lines correspond to different hyperparamters under the same strategy. 
Falling closer to the origin of the graph indicates higher efficiency in accuracy-latency trade-off.
%\textbf{differnet data points are under the same strategies. Closer to origin the better.}
%Detailed numerical results for the key data points are in Table \ref{tab.full}.
%\input{figs/tradeoff}
\begin{table*}[t]
	\footnotesize
	\caption{Incremental ASR performance on How2 300h English before and after adaptation. Latency is reported as difference to the unadapted model under the hold-$0$ strategy. In parentheses are changes in WER after adaptation.}
	\centering
%	\resizebox{0.45\textwidth}{!}{
		\begin{tabular}{p{1.9cm}rrrrrrrrrrrrrrr}
			\toprule
			& \multicolumn{4}{c}{\textbf{Unadapted}} && \multicolumn{4}{c}{\textbf{Adapted to Partial Inputs}} \\
			\textbf{Strategies} & \multicolumn{2}{c}{\textbf{WER (\%)} $\downarrow$} & \multicolumn{2}{c}{\textbf{$\Delta$ latency (sec.)} $\downarrow$} && \multicolumn{2}{c}{\textbf{WER (\%)} $\downarrow$} & \multicolumn{2}{c}{\textbf{$\Delta$ latency (sec.)} $\downarrow$}   \\
			& \makecell[c]{Bidir.} & 
			\makecell[c]{Unidir.} & 
			\makecell[c]{Bidir.} &
			\makecell[c]{Unidir.} &&
			\makecell[c]{Bidir.} & 
			\makecell[c]{Unidir.} & 
			\makecell[c]{Bidir.} &
			\makecell[c]{Unidir.} \\%& 
%			\makecell[c]{BLEU} & \makecell[c]{$\Delta$ latency (sec.)} & \\%& \makecell[c]{BLEU (\%)} & \makecell[c]{$\Delta$ latency (sec.)} && 			\makecell[c]{BLEU (\%)} & \makecell[c]{$\Delta$ latency (sec.)} \\
			\midrule
			Hold-0 & 37.4 & 40.7 & 0 & 0 && 28.3 ($-$9.1) & 30.3 ($-$10.4) & $+$0.32 & $+$0.30 \\%& \textbf{37.4}  &  0    & & 41.3 &  0    && 40.7 &  0    \\	
			\midrule
%			\multicolumn{1}{l}{\textbf{Hold-$n$}} & \\
			Hold-2 & 20.4 & 22.0 & $+$0.48 & $+$0.42 && 18.7 ($-$1.7) & 19.3 ($-$2.7)  & $+$0.77 & $+$0.73 \\			
			Hold-4 & 17.8 & 18.8 & $+$0.92 & $+$0.83 && 17.1 ($-$0.7) & 17.3 ($-$1.5) & $+$1.22 & $+$1.16 \\
			Hold-6 & 16.8 & 17.6 & $+$1.34 & $+$1.22 && 16.5 ($-$0.3) & 16.5 ($-$1.1)  & $+$1.60 & $+$1.52\\
			%%%%%%%%%%%%%%%%%%%%%%%%%%%%%%
%			\multicolumn{1}{l}{\textbf{Output-if-agree}} &\\	
			Local agreement & 15.8 & 16.8 & $+$0.81 & $+$0.54 && 15.8 ($-$0.0) & \textbf{15.5} ($-$1.0) & $+$0.83 & $+$\textbf{0.65} \\	
			%%%%%%%%%%%%%%%%%%%%%%%%%%%%%%
			\midrule
			Offline & 14.9 & 14.4 & $+$4.55 & $+$4.47 && 14.4 ($-$0.5) & 14.7 ($+$0.3) & $+$4.59 & $+$4.45 \\%& \textcolor{oCl}{14.9} & \textcolor{oCl}{+4.55} && \textcolor{oCl}{14.2} &     \textcolor{oCl}{+4.31} && \textcolor{oCl}{14.4} & \textcolor{oCl}{+4.47} \\
			\bottomrule
		\end{tabular}
%	}
	\label{tab.adapt}
\end{table*}

The first observation from Figure \ref{fig.tradeoffHow2} is that the local agreement strategy outperforms other strategies by a large margin.
%Based on determining its output based on agreements between consecutive chunks 
Reducing latency by 3.8 seconds at the cost of 0.9\% WER compared to offline system, it achieves the most efficient accuracy-latency trade-off.
It shows that dynamically selecting partial hypotheses is preferable to hard thresholds set by hyperparameters.
%Compared to the offline system with 14.9\% and roughly 4.6 seconds latency, losing 0.9\% while gaining dramatically reducing latency is profitable.
%% analyze wait-k
Comparing the other two strategies, initially surprising to us was that hold-$n$ consistently outperforms wait-$k$, a proven method in simultaneous translation.
A further analysis shows that this method suffers from varying utterance speed.
In text translation, the input-output rate is roughly 1:1. 
In ASR, however, a fixed speed for all chunks is a too strong assumption.
 %seems harmful to finding an optimal trade-off between accuracy and latency.
Indeed, at a fixed output rate of $r=2$, we get close to offline WER (15.9\% vs 14.9\%). %, especially in the configurations that wait for the initial chunk (``+ rate 2"). 
However, this comes with large latency of nearly 2.5 seconds.
As 2 tokens per second corresponds to a very low utterance speed, this output rate leads to most of the tokens being displayed after the utterance ends, in effect converging towards an offline system.
On the other hand, increasing the output rate quickly results in more errors.
Given a high output rate, the allowed number of output tokens is likely to exceed the actual number of tokens said in the chunk. 
In this case, we approach the other extreme, the hold-$0$ strategy that outputs all chunk-level tokens.
Given this finding, we eliminate this option and only pursue the hold-$n$ and the local agreement strategies in the upcoming experiments. 

%Overall, with the current set of strategies, there is still a gap of 1\%. 

%Comparing the strategies: \textbf{if-agree} is best. hold-n is consistently better than wait-k.

%% overall
%Despite the difficulty of finding a suitable output rate, it is interesting that waiting consistently brings large performance gains over the counterparts without waiting. 
%This provides evidence for our hypothesis that the output quality in the earlier chunks tend to be low due to the lack of input context. 
%Conditioning future outputs with these outputs in full is likely to lead to low-quality transcriptions.

%\input{tabs/offline}
%Table \ref{tab.bounds}

%%%%%%%%%%%%%%%%%%%%%%%%%%%%%%%%%%%%%%%%%%%%%%%%%%%%%%%%%%%%%%%%%%%%%%%%%%%%%%%

\subsection{Low-Latency ASR}
After identifying promising hypothesis selection strategies, we proceed to incremental inference.
We first focus on the ASR performance of the unidirectional model. 
Then we study the effect of adaptation to partial sequences.

As shown in the bottom left section of Table \ref{tab.adapt}, the unidirectional full-sequence model achieves similar accuracy to the bidirectional one (WER 14.4\% vs 14.9\%).
However, the incremental inference WER with the unidirectional model is constantly around 1\% higher than the bidirectional model.
This suggests that the unidirectional model, when directly performing online inference, is more susceptible to the loss of future context.
However, the gap closes after adaptation on partial sequences.
The unidirectional model becomes on-par with its bidirectional counterpart under the more promising strategies, e.g. hold-4 and local agreement.
Furthermore noteworthy is that after adaptation it still preserves the performance under offline inference, as evidenced by the WER of 14.4\% and 14.7\% before and after adaptation.

Contrasting the right-hand-side section of Table \ref{tab.adapt} with the left, we see the effect of adaptation.
%The same strategy before and after adaptation is plotted with the same number of data points. 
%Based on a point’s order in the interpolated line, we can locate its (un)adapted counterpart. 
The error reduction after adaptation comes with a slight increase in latency up to 0.3 second.
%This suggests that adaptation helps error reduction at the expense of a minor increase in latency.
It is a sign that the model can better control its chunk-level outputs.
Moreover, the impact is stronger with the unidirectional model.
A hypothesized reason is that its encoder representation already has no dependency on future inputs, adaptation to partial sequences is therefore an easier task.

%\input{figs/adapt}
%Overall, it is clear that adaptation significantly reduces WER. 
%Here, the effect is stronger with the unidirectional case, as shown by \textbf{numbers!!!!!}
%\textbf{why is stronger?}

%\input{figs/tradeoff2}

%Now, both the baseline and the unidirectional model are fine-tuned with partial sequences.
%Before considering online decoding, we need to verify whether the adaptation to partial inputs negatively impacted the performance on full sequences. 
%As shown by the offline accuracy in the bottom row for each strategy, adaptation does not significant change WER. 
%This confirms that fine-tuning on partial sequences does not harm offline prediction performance.

\subsection{Low-Latency Speech Translation}
Having seen the effectiveness of the unidirectional model in low-latency ASR, we validate the findings on speech translation, a different sequence-to-sequence task.
The model here is before partial sequence adaptation. %, although further gains are possible. 
% since the input and output alignment is not necessarily monotonic.

Table \ref{tab.stl} outlines the performance and latency of the unidirectional model on the How2 English-Portuguese translation task.
We use BLEU \cite{papineni-etal-2002-bleu, post-2018-call} and METEOR \cite{banerjee-lavie-2005-meteor} as quality metrics.
The results here agree with the previous findings on ASR. 
In general, by selectively taking prefixes of chunk-level hypotheses, we can largely reduce latency by sacrificing some output quality.
More importantly, the local agreement selection strategy remains to achieve the most efficient quality-latency trade-off.
It scores similarly to the hold-4 strategy in term of the quality metrics but has 0.2 second less latency.
Compared to the offline system, we see an 84\% relative reduction in latency with a 5\% relative loss in BLEU and METEOR.

\begin{table}[h]
	\footnotesize
	\caption{Incremental speech translation performance on How2 300h English-Portuguese.}
	\centering
	\resizebox{0.45\textwidth}{!}{
		\begin{tabular}{p{1.9cm}rrrrrrrrrrrrrrrrrr}
			\toprule
			\textbf{Strategies} & 
			\makecell[c]{\textbf{BLEU}\footnotemark $\uparrow$\\} & \makecell[c]{\textbf{METEOR} $\uparrow$} & \makecell[c]{\textbf{$\Delta$ latency (sec.) $\downarrow$}} \\%& 
%			\makecell[c]{BLEU} & \makecell[c]{$\Delta$ latency (sec.)} & \\%& \makecell[c]{BLEU (\%)} & \makecell[c]{$\Delta$ latency (sec.)} && 			\makecell[c]{BLEU (\%)} & \makecell[c]{$\Delta$ latency (sec.)} \\
			\midrule
			Hold-0 & 24.9  & 25.5 & 0 \\%& \textbf{37.4}  &  0    & & 41.3 &  0    && 40.7 &  0    \\	
			\midrule
%			\multicolumn{1}{l}{\textbf{Hold-$n$}} & \\
			Hold-2 & 37.3  & 31.4 & $+$0.48 \\			
			Hold-4 & 42.2  & 33.6 & $+$0.95 \\
			Hold-6 & 43.6  & 34.2 & $+$1.38 \\
			%%%%%%%%%%%%%%%%%%%%%%%%%%%%%%
%			\multicolumn{1}{l}{\textbf{Output-if-agree}} &\\	
			Local agreement & 42.1 & 33.5 &  $+$0.71 \\	
			%%%%%%%%%%%%%%%%%%%%%%%%%%%%%%
			\midrule
			Offline & 44.5 & 34.5  & $+$4.36 \\%& \textcolor{oCl}{14.9} & \textcolor{oCl}{+4.55} && \textcolor{oCl}{14.2} &     \textcolor{oCl}{+4.31} && \textcolor{oCl}{14.4} & \textcolor{oCl}{+4.47} \\
			\bottomrule
		\end{tabular}
	}
	\label{tab.stl}
\end{table}
\footnotetext{SacreBLEU: case.mixed+numrefs.1+smooth.exp+tok.13a+version.1.4.3}

\section{Conclusion}
In this paper, we explored approaches for latency reduction in sequence-to-sequence speech recognition and translation.
First, we studied the accuracy-latency trade-off under chunk-based incremental inference. 
By selecting a subset of chunk-level outputs, we achieved large error reduction with minimal delay.
Among the three partial hypothesis selection strategies, the most efficient trade-off was achieved by examining the local agreement of hypotheses.
To facilitate incremental inference, we trained a unidirectional model where the encoder accesses no future context.
With a simple yet effective adaptation procedure on partial inputs, the unidirectional model performed on-par with its bidirectional counterpart on ASR.
On the 300-hour How2 dataset, we were able to reduce the gap to offline systems to less than 1\% absolute while incurring 0.7 seconds of latency.
Besides ASR, our approach was also effective on speech translation.
A future direction is to incorporate adaptive chunk sizes.

\iffalse 
\begin{itemize}
\item first, chunk-based decoding strategies: Show trade-off. Find promising output-selection strategies. Can hugely reduce latency by a small sacrifice of accuracy, x\% to offline situation with latency of 1 second.
\item unidirectional: we remove the encoder’s dependency on future inputs. This way, as a sequence incrementally streams in, only the new inputs need to be encoded. Performance similar to bidirectional but still a gap (despite the computational advantage)
\item adaptation to partial sequences. Makes more efficient exchange between latency and accuracy. 
After adaptation, unidirectional ~~ baseline shrank to minimal.
We furthermore found that adaptation enables a more efficient accuracy latency trade-off in general, regardless of strategy.
With the strongest-performance strategy, we can nearly match the offline accuracy with a latency of within 1 second.
\item We show that the approach is also applicable beyond transcription - also to spoken language translation.
\item future work: chunk size?
\end{itemize}
\fi

\section{Acknowledgment}
The experiments were supported by the Google Cloud Platform research credit program.

\newpage
\bibliographystyle{IEEEtran}
\bibliography{mybib}

% \begin{thebibliography}{9}
% \bibitem[1]{Davis80-COP}
%   S.\ B.\ Davis and P.\ Mermelstein,
%   ``Comparison of parametric representation for monosyllabic word recognition in continuously spoken sentences,''
%   \textit{IEEE Transactions on Acoustics, Speech and Signal Processing}, vol.~28, no.~4, pp.~357--366, 1980.
% \bibitem[2]{Rabiner89-ATO}
%   L.\ R.\ Rabiner,
%   ``A tutorial on hidden Markov models and selected applications in speech recognition,''
%   \textit{Proceedings of the IEEE}, vol.~77, no.~2, pp.~257-286, 1989.
% \bibitem[3]{Hastie09-TEO}
%   T.\ Hastie, R.\ Tibshirani, and J.\ Friedman,
%   \textit{The Elements of Statistical Learning -- Data Mining, Inference, and Prediction}.
%   New York: Springer, 2009.
% \bibitem[4]{YourName17-XXX}
%   F.\ Lastname1, F.\ Lastname2, and F.\ Lastname3,
%   ``Title of your INTERSPEECH 2020 publication,''
%   in \textit{Interspeech 2020 -- 20\textsuperscript{th} Annual Conference of the International Speech Communication Association, September 15-19, Graz, Austria, Proceedings, Proceedings}, 2020, pp.~100--104.
% \end{thebibliography}

\end{document}